\documentclass{article} 
\usepackage[preprint]{colm2026_conference}

\usepackage{microtype}
\usepackage{hyperref}
\usepackage{url}
\usepackage{multirow}
\usepackage{booktabs}
\usepackage{graphicx}
\usepackage{subcaption}
\usepackage{amsmath}
\usepackage{algorithm}
\usepackage{algpseudocode}
\usepackage{xcolor}
\usepackage[table]{xcolor}
\usepackage{float}
\newcommand{\algcomment}[1]{\hfill {\color{gray}#1}}
\definecolor{mycolor}{RGB}{230,220,250}

\usepackage{lineno}

\definecolor{darkblue}{rgb}{0, 0, 0.5}
\hypersetup{colorlinks=true, citecolor=darkblue, linkcolor=darkblue, urlcolor=darkblue}

\title{OASIS: Online Activation Subspace Learning for Memory-Efficient Training}

\author{Sakshi Choudhary$^1$, Utkarsh Saxena\thanks{Work done while at Purdue University.}   \hspace{0.5pt} $^{2}$ , Kaushik Roy$^1$ 
\\
$^1$Purdue University, $^2$Advanced Micro Devices, Inc. (AMD)\\
}

\begin{document}

\ifcolmsubmission
\linenumbers
\fi

\maketitle
\begin{abstract}
Training large language models (LLMs) is constrained by memory requirements, with activations accounting for a substantial fraction of the total footprint. Existing approaches reduce memory using low-rank weight parameterizations or low-rank gradient subspaces for optimizer states, while activation memory is addressed through architectural modifications or compression schemes based on periodically updated projections.
We propose OASIS, an online activation subspace learning algorithm for memory-efficient training that tracks and continuously updates a low-dimensional activation subspace during training. Intermediate activations are projected onto this evolving subspace, reducing memory without modifying forward-pass computations. 
The evolving activation subspace induces low-rank gradient representations, enabling both gradients and optimizer states to be maintained directly in this subspace, while a projection-aware optimizer consistently transports optimizer states across subspace updates for stable training.
Across various finetuning and pretraining tasks, OASIS achieves up to $2\times$ lower peak memory than full fine-tuning while matching its performance and outperforming prior low-rank methods. \footnote{https://github.com/Sakshi09Ch/OASIS}
\end{abstract}
\section{Introduction}

\begin{figure}[t]
\centering
\begin{subfigure}[t]{0.45\linewidth}
\centering
\includegraphics[width=\textwidth]{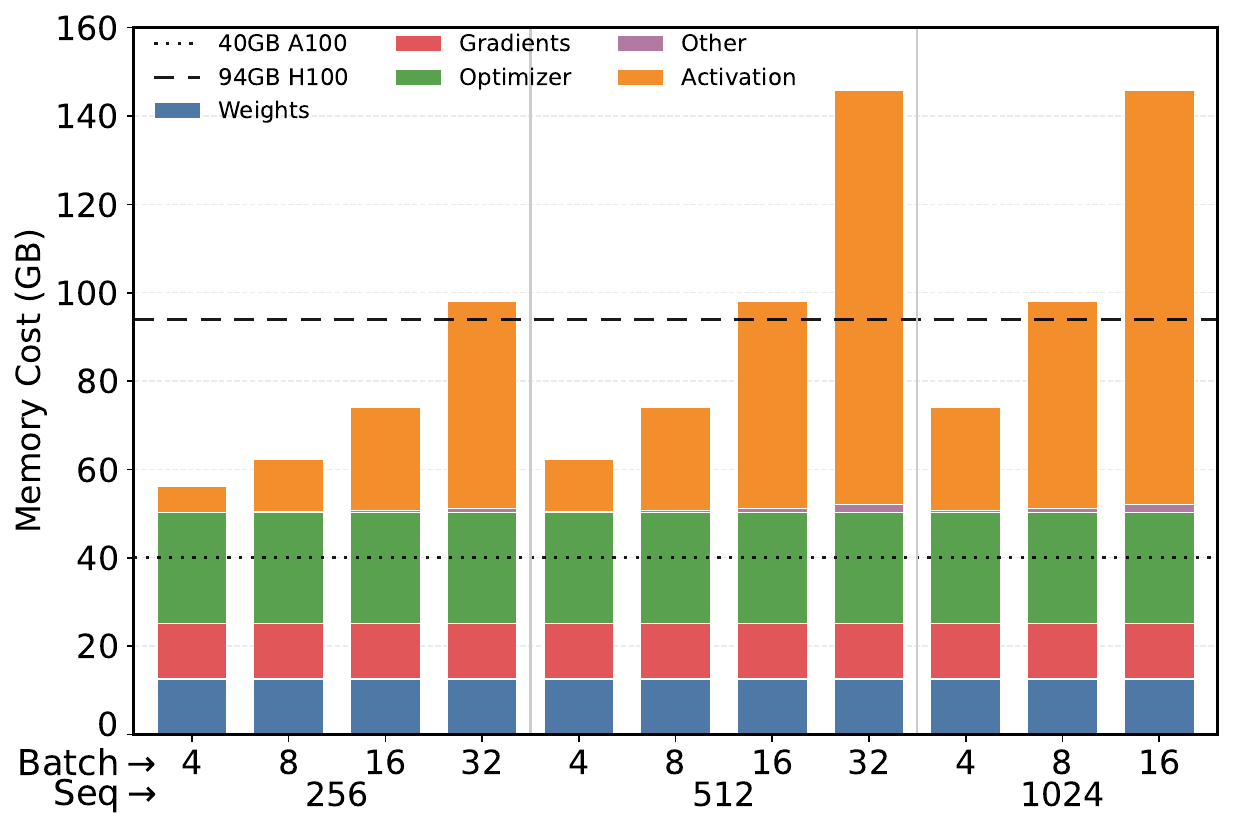}
\caption{Memory breakdown during training}\label{fig:intro_a}
\end{subfigure}
\hfill
\begin{subfigure}[t]{0.45\linewidth}
\centering
\includegraphics[width=\textwidth]{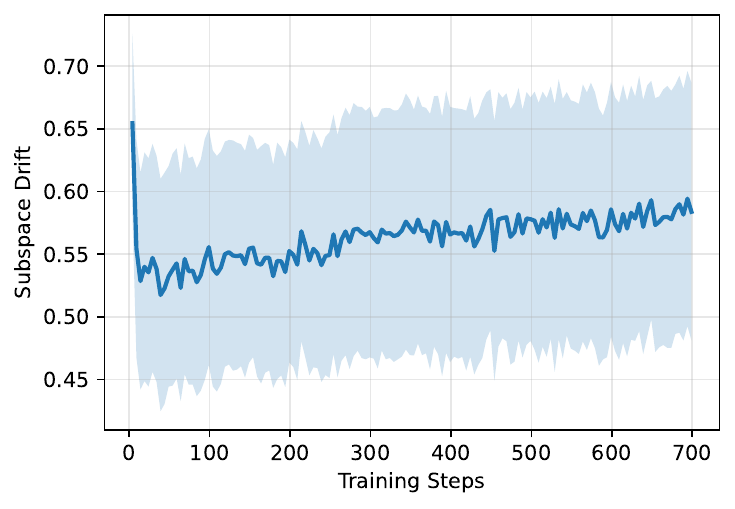}
\caption{Activation subspace drift}\label{fig:intro_b}
\end{subfigure}
\caption{
Memory breakdown and subspace dynamics during training. (a) As batch size and sequence length scale, memory becomes increasingly dominated by activations.
(b) These activations lie in subspaces that evolve continuously during training.
}
\label{fig:intro_motivation}
\vspace{-2.5mm}
\end{figure}

Training large language models (LLMs) is increasingly constrained by memory rather than compute alone \citep{zero,nvidia_seq}. In addition to model parameters and gradients, training requires storing optimizer states, which typically increase parameter memory by approximately 3× \citep{adam}, as well as intermediate activations for backpropagation. These costs become especially severe in large-scale training regimes, where activation memory scales with batch size and sequence length and optimizer-state memory grows with model size, together dominating the total training footprint, as illustrated in Figure~\ref{fig:intro_a}.

A promising direction is to exploit the low-rank structure of weights, gradients, and activations during training. Prior work has leveraged low-rank parameterizations \citep{lora} and gradient subspaces \citep {galore,ldadam} to reduce optimizer memory while maintaining full-parameter updates. However, these approaches do not reduce the memory required to store activations for backpropagation. Conversely, methods that target activation memory footprint often require architectural changes \citep{cola} or introduce approximation errors that can lead to suboptimal performance \citep{velora}. Consequently, activation memory, often the dominant component at scale, remains inadequately addressed.


In this paper, we propose OASIS, an activation-aware low-rank training algorithm based on a simple but powerful observation: for a linear layer, the weight gradients lie in the span of input activations \citep{actspan}. This motivates identifying a low-rank activation subspace that can be used to represent both input activations and the gradients.
Gradients inherit the low-rank structure induced by this subspace and can therefore be represented in low rank, reducing gradient memory and naturally inducing low-rank optimizer states. Unlike prior approaches that target individual components of training memory, our method compresses activations, gradients, and optimizer states through a single algorithm.

In practice, this subspace is not static. As model parameters evolve, the distribution of activations shifts, causing the underlying low-rank subspace to drift over time. To demonstrate this, we measure the average activation subspace drift across all layers between successive training iterations during Llama 2 7B \citep{llama2} finetuning on GSM8K \citep{gsm8k}, with higher values indicating greater drift. As shown in Figure \ref{fig:intro_b}, the drift remains consistently non-zero throughout training, indicating that the activation subspace continues to evolve rather than stabilizing to a fixed basis. A fixed projection basis therefore becomes progressively stale and can degrade optimization.
To address this, we develop an online activation subspace learning mechanism based on Oja's rule \citep{oja}, which incrementally updates the low-rank subspace during training without requiring repeated exact eigen-decomposition. As the projection basis evolves, the compressed optimizer states (e.g., momentum and variance) must also be updated consistently. Following LDAdam \citep{ldadam}, we use projection-aware updates to ensure alignment with the current subspace and maintain stable low-rank optimization.

We evaluate OASIS on both finetuning and pretraining tasks across a range of model scales. Experiments on GSM8K \citep{gsm8k}, HumanEval \citep{humaneval}, and C4 \citep{C4} with Llama models show that OASIS achieves substantial memory savings while maintaining competitive performance. It reduces peak memory by up to $2\times$ compared to full fine-tuning and over $30\%$ relative to LDAdam \citep{ldadam} at comparable accuracy, while outperforming prior low-rank methods in aggressive compression regimes. These gains persist in pretraining, demonstrating effectiveness across model scales and training settings.
Our contributions are as follows:

\begin{itemize}
    \item We introduce OASIS, an online activation subspace learning algorithm for memory-efficient training, which leverages the low-rank structure of activations to jointly compress stored activations, gradients, and optimizer states.

    \item We develop an online subspace adaptation mechanism based on Oja’s rule that incrementally tracks the evolving activation subspace during training without requiring repeated exact eigen decomposition on activation statistics.

    \item Through extensive experiments on finetuning and pretraining tasks, we show that OASIS achieves up to $2\times$ lower peak memory than full fine-tuning while consistently outperforming prior low-rank methods.
\end{itemize}

\section{Related Works}
\textbf{Low-Rank Parameterization.} Low-rank parameterization methods such as LoRA \citep{lora} and its variants \citep{adalora,qlora,dora} restrict updates to trainable low-rank adapters while keeping the backbone weights frozen, thus significantly reducing the number of trainable parameters and the memory footprint associated with gradients and optimizer states. However, these approaches restrict optimization to a fixed low-dimensional update space, which can limit their ability to match full-parameter training performance \citep{galore,loralearnsless}. To address this limitation, subsequent methods increase the effective rank of updates by aggregating multiple low-rank updates over time \citep{relora,chainoflora}. However, they typically require an initial warm-up phase with full-parameter training \citep{relora}, which undermines the benefits of low-rank optimization.

\textbf{Low-Rank Optimization via Gradient Subspaces.} A complementary line of work performs full-parameter optimization while compressing optimizer states by exploiting the low-rank structure of gradients \citep{galore, ldadam, microadam, subtrack, osd}. GaLore \citep{galore} periodically applies Singular Value Decomposition (SVD) to identify dominant gradient directions and construct projection matrices defining a low-dimensional gradient subspace. Gradients are projected onto this subspace to update optimizer states, reducing memory while preserving full-parameter updates.
LDAdam \citep{ldadam} improves upon this by introducing error-feedback and projection-aware state updates to maintain consistency as the subspace evolves. Subsequent works reduce the overhead of periodic SVD through randomized projections \citep{flora} or streaming subspace estimation via online updates \citep{osd,subtrack}.
However, these methods focus on reducing optimizer-state memory and do not scale to large-scale training settings where activation memory dominates the model footprint (Figure~\ref{fig:intro_a}).

\textbf{Activation Compression during Training.} A common approach to reducing activation memory is to avoid storing intermediate activations and instead recompute them on-the-fly during the backward pass \citep{actckpt}. While effective, this shifts the burden from memory to compute, resulting in increased training time.
VeLoRA \citep{velora} reduces activation memory by splitting tokens into sub-tokens and compressing them into a one-dimensional subspace. However, it requires projecting activations back to full space for gradient computation and is used only for a subset of layers, limiting its generality.
CompAct \citep{compact} applies low-rank compression to stored activations using random projections, which can slightly reduce the computational cost of GaLore-style optimization, but at the expense of accuracy, underperforming both GaLore and full-parameter training. Similarly, LANCE \citep{lance} utilizes SVD to obtain low-rank activation subspace for efficient on-device continual learning setup and is not directly applicable to standard training.
Taking a different approach, CoLA \citep{cola} modifies the model architecture by replacing linear and projection layers with autoencoders to induce low-rank activations, but this restricts its applicability to fine-tuning existing LLMs. Overall, these approaches either require invasive architectural modifications or achieve memory savings at the cost of performance, limiting their effectiveness for general-purpose training.
Please refer to Appendix~\ref{sec:extendrelated} for additional discussion on activation compression for efficient inference.

\section{Methodology}

OASIS reduces the memory footprint of training large models by performing optimization in a dynamically evolving low-dimensional subspace of activations. The key idea is to compress the activations stored for the backward pass while keeping the forward pass exact, enabling memory savings without altering forward computations.
Consider a linear layer with input activations $X_t \in \mathbb{R}^{N \times d}$ at iteration $t$, and output gradients $G_t^{\text{out}} \in \mathbb{R}^{N \times m}$. The gradient with respect to the weight matrix can be written as
\begin{equation}
\nabla_W = X_t^\top G_t^{\text{out}},
\end{equation}
which lies in the span of the input activations. This enables low-rank optimization by projecting activations onto a rank-$r$ subspace using an orthonormal basis $U_t \in \mathbb{R}^{d \times r}$:
\begin{equation}
\widetilde{X}_t = X_t U_t.
\end{equation}
The resulting low-rank gradient is
\begin{equation}
\widetilde{G}_t = \widetilde{X}_t^\top G_t^{\text{out}},
\end{equation}
corresponding to optimal rank-$r$ approximation of the gradient within the activation span.

This formulation avoids materializing full-rank gradients and allows both gradients and optimizer states to be maintained directly in the low-rank subspace. For optimizers such as Adam, the first- and second-moment estimates are computed and stored in $\mathbb{R}^{r \times m}$ using $\widetilde{G}_t$, rather than $\mathbb{R}^{d \times m}$. Parameter updates computed using $\widetilde{G}_t$ are projected back to the full space via $U_t$. This reduces activation storage from $O(Nd)$ to $O(Nr)$ and both gradient and optimizer state storage from $O(dm)$ to $O(rm)$ per layer.

To remain effective, this approach requires maintaining a subspace $U_t$ that captures the dominant directions of the activation distribution throughout training. We initialize $U_0$ using principal components of early activations and update it online using an efficient streaming method described next. Algorithm~\ref{alg:oasis} summarizes the overall procedure.

\subsection{Online Activation Subspace Learning}
A key challenge in subspace-based optimization is maintaining a basis $U_t$ that tracks the evolving activation subspace during training. As model parameters are updated, the activation distribution changes, and the optimal low-rank subspace is inherently time-varying. This requires a method that can adapt the subspace continuously and efficiently.

Existing approaches typically rely on periodic subspace updates via Singular Value Decomposition (SVD) \citep{galore} or on per-iteration approximations computed from individual mini-batches \citep{ldadam}. Periodic updates can lead to stale subspaces between updates, while batch-level methods are often noisy and fail to capture the temporal evolution of the activation distribution. As a result, both approaches can yield suboptimal projections and degrade optimization performance.

To address this, we adopt an online subspace update based on Oja’s rule \citep{oja}, a classical streaming PCA method designed to estimate principal components from sequential data. This makes it well-suited for training settings, where data arrive in mini-batches and the underlying subspace evolves over time. The update incrementally steers the basis toward directions of high variance while removing components already captured by the current subspace:
\begin{equation}
U_t \leftarrow U_{t-1} + \gamma (I - U_{t-1} U_{t-1}^\top) C_t U_{t-1},
\end{equation}
where $C_t = \frac{1}{N} X_t^\top X_t$ denotes the activation covariance at iteration $t$.

Here, $C_t U_{t-1}$ biases the basis toward the principal directions of the activation distribution, while the projection $(I - U_{t-1} U_{t-1}^\top)$ removes components already captured by the current subspace. This yields an efficient incremental approximation to principal component analysis without requiring explicit eigendecomposition.

However, directly applying this update introduces two challenges. First, the update does not preserve orthonormality, and repeated iterations can cause the basis vectors to become correlated. Maintaining an orthonormal basis is essential to ensure that projections correspond to least-squares optimal approximations. Second, the effective update magnitude depends on the scale of $C_t$, which varies across layers and training iterations. A fixed step size can therefore lead to instability when activation magnitudes are large, or slow adaptation when they are small.

To address these issues, we make two modifications. We explicitly re-orthonormalize $U_t$ after each update to maintain an orthonormal basis, incurring a cost of $O(dr^2)$, which is significantly cheaper than full eigendecomposition. We also normalize the step size using
\begin{equation}
\gamma_t = \frac{\gamma}{\|C_t\|},
\end{equation}
making the update adaptive to activation scale and improving stability across layers and training stages.
Additionally, we adopt a projection-aware adaptive optimizer inspired by \citet{ldadam} to ensure consistency under subspace evolution.

\begin{algorithm}[H]
\caption{OASIS}
\label{alg:oasis}
\textbf{Notation:} Step size $\eta_t$, decay rates $\beta_1,\beta_2$, Oja step size $\gamma$, rank $r$. 
$W \in \mathbb{R}^{d \times m}$ denotes the weight matrix of a given layer.
$X_t \in \mathbb{R}^{N \times d}$ and $G_t^{\text{out}} \in \mathbb{R}^{N \times m}$ are the corresponding input activations and output gradients, where $N = bn$, with $b$ the batch size and $n$ the sequence length. $T_t = U_t^\top U_{t-1}$ represents the subspace transition matrix

\textbf{Initialization:} Use $X_0$ to form $C_0 = \frac{1}{N}X_0^\top X_0$ and initialize
$U_0 \in \mathbb{R}^{d \times r}$ as the top-$r$ principal components.
Set $M_0 = 0$ and $V_0 = 0$. $|\cdot|$ denotes element-wise absolute value.

\begin{algorithmic}[1]
\For{$t = 1,2,\dots,T$}
    \State $C_t \gets \frac{1}{N}X_t^\top X_t$ \algcomment{//Activation covariance}
    \State $U_t \leftarrow U_{t-1} + \frac{\gamma}{\|C_t\|} (I - U_{t-1} U_{t-1}^\top) C_t U_{t-1}$ \algcomment{//Online subspace update (Oja-based)}
    \State $U_t \gets \mathrm{orth}(U_t)$ \algcomment{//Re-orthonormalize basis}
    \State $\widetilde{X}_t \gets X_t U_t$ 
    \State $\widetilde{G}_t \gets \widetilde{X}_t^\top G_t^{\text{out}}$ 
    \State $M_t \gets \beta_1 (T_t M_{t-1}) + (1-\beta_1)\widetilde{G}_t$ \algcomment{//Projection aware optimizer}
    \State $V_t \gets \beta_2 \Big[(1-\beta_2^{t-1})\Big|(T_t)^2 \odot (V_{t-1}-M_{t-1}^2) + (T_tM_{t-1})^2\Big|\Big] + (1-\beta_2)\widetilde{G}_t^2$ 
    \State $\widehat{M}_t \gets M_t / (1-\beta_1^t)$ ; $\widehat{V}_t \gets V_t / (1-\beta_2^t)$
    \State $\widetilde{N}_t \gets \widehat{M}_t / (\sqrt{\widehat{V}_t} + \epsilon)$ 
    \State $\widehat{G}_t \gets U_t \widetilde{N}_t$ \algcomment{// Map update to full space}
    \State $W_{t+1} \gets W_t - \eta_t \widehat{G}_t$ \algcomment{// Parameter update}
\EndFor
\end{algorithmic}
\end{algorithm}

\subsection{Projection Aware Optimizer}\label{sec:projaware}
We maintain optimizer states in a low-dimensional activation subspace that evolves during training. As the subspace changes, optimizer states are transported to remain aligned with the current basis. Let $U_{t-1}$ and $U_t$ denote the orthonormal bases at iterations $t-1$ and $t$. The subspace transition matrix $T_t = U_t^\top U_{t-1}$. defines the change of coordinates between subspaces and is used to transport optimizer states from one low-rank subspace to another.

Following prior work~\citep{ldadam}, we view Adam's states as coordinate-wise estimates of gradient moments. Under this view, the first moment $M_t$ can be directly projected between subspaces using the linear transformation induced by $T_t$ (Algorithm~\ref{alg:oasis}, Line~7). The second moment $V_t$, however, cannot be directly projected between subspaces, as it depends on cross-coordinate interactions not captured by coordinate-wise estimates. We approximate it in the new subspace using projected variance and mean-squared terms derived from the transported moments, avoiding explicit covariance estimation (Algorithm~\ref{alg:oasis}, Line~8).

\section{Experiments}
\subsection{Experimental Setup}
\textbf{Models and Datasets.}
We evaluate OASIS across both finetuning and pretraining using language models of varying scales. For downstream tasks, we use Llama-2 7B \citep{llama2} and Llama-3.2 1B \citep{llama3}, finetuned on the GSM8K dataset \citep{gsm8k}. For code generation, we finetune Llama-2 7B on CodeAlpaca \citep{codealpaca} and evaluate on HumanEval \citep{humaneval}. To study the behavior of OASIS in the pretraining regime, we train Llama-130M and Llama-350M from scratch on the C4 dataset \citep{C4}.

\textbf{Training Details.} All models are trained using the Adam optimizer. We report task-specific performance metrics: accuracy on GSM8K, pass@10 on HumanEval, and validation loss on C4. In addition, we measure peak memory usage during training to quantify the efficiency gains of OASIS.  Please refer to Appendix \ref{apex:hyper} for detailed hyperparameters

\textbf{Baselines.} We compare OASIS against diverse training strategies: (i) \textbf{Full fine-tuning} with Adam, which serves as the standard upper bound, (ii) \textbf{LoRA}~\citep{lora}, a parameter-efficient finetuning method that updates low-rank adapters while freezing the base model, (iii) \textbf{GaLore}~\citep{galore}, which reduces the memory footprint of optimizer states by projecting gradients into a low-rank subspace, and (iv) \textbf{LDAdam}~\citep{ldadam}, which similarly operates in a low-dimensional gradient subspace while incorporating error feedback and projection-aware optimizer to improve performance.

\begin{table}[t]
\centering
\begin{tabular}{l l l cc}
\toprule
Model & Rank & Method & Accuracy (\%) & Peak Memory (GB) \\
\midrule

\multirow{9}{*}{Llama-2 7B}
 & -- & Adam & 39.37 $\pm$ 0.39 & 95.18 \\

\cmidrule(lr){2-5}

 & \multirow{4}{*}{32}
 & LoRA   & 37.13 $\pm$ 0.64 & 64.15 \\
 &       & GaLore & 35.96 $\pm$ 1.18 & 71.30 \\
 &       & LDAdam & 38.74 $\pm$ 0.11 &  71.48\\
 &       & \cellcolor{mycolor}OASIS & \cellcolor{mycolor}39.14 $\pm$ 0.72 & \cellcolor{mycolor}48.21 \\

\cmidrule(lr){2-5}

 & \multirow{4}{*}{128}
 & LoRA   & 37.27 $\pm$ 0.36 & 66.00 \\
 &       & GaLore & 37.49 $\pm$ 0.56 & 72.03 \\
 &       & LDAdam & 38.17 $\pm$ 0.86 &  72.21\\
 &       & \cellcolor{mycolor}OASIS & \cellcolor{mycolor}39.10 $\pm$ 1.27 &  \cellcolor{mycolor}49.60\\

\midrule
\addlinespace

\multirow{9}{*}{Llama-3 1B}
 & -- & Adam & 27.09 $\pm$ 0.58 & 52.54 \\

\cmidrule(lr){2-5}

 & \multirow{4}{*}{32}
 & LoRA   & 22.51 $\pm$ 0.46 & 48.06 \\
 &       & GaLore & 21.64 $\pm$ 0.36 & 47.17 \\
 &       & LDAdam & 24.55 $\pm$ 0.91 & 49.01 \\
 &       & \cellcolor{mycolor}OASIS  & \cellcolor{mycolor}23.78 $\pm$ 0.26 & \cellcolor{mycolor} 41.03 \\

\cmidrule(lr){2-5}

 & \multirow{4}{*}{128}
 & LoRA   &  26.61 $\pm$ 0.62 & 48.62 \\
 &       & GaLore & 25.51 $\pm$ 0.61 & 47.40 \\
 &       & LDAdam &  25.72 $\pm$ 1.07 & 49.23 \\
 &       & \cellcolor{mycolor}OASIS & \cellcolor{mycolor}26.88 $\pm$ 0.49 & \cellcolor{mycolor}41.59 \\

\bottomrule
\end{tabular}
\caption{Finetuning results on GSM8K for Llama-2 7B and Llama-3 1B. For Llama-2 7B, OASIS matches the performance of Adam while reducing peak memory by $\sim2\times$.}
\label{tab:finetune_llama_gsm8k}
\end{table}

\begin{figure}[t]
\centering
\includegraphics[width=0.7\textwidth]{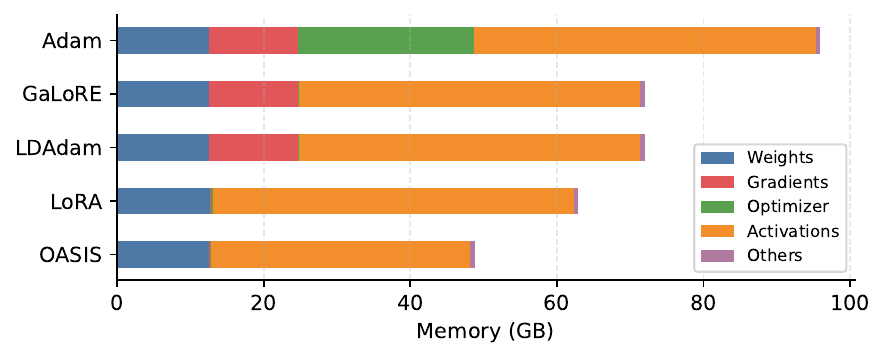}
\caption{Memory breakdown across training components for different methods on LLaMA-2 7B GSM8K finetuning at rank 32. OASIS reduces memory across activations, gradients, and optimizer states, leading to the lowest overall memory footprint.}
\label{fig:memory_breakdown_baselines}
\end{figure}

\begin{table}[ht]
\centering
\begin{tabular}{l l l cc}
\toprule
Rank & Method & Accuracy (\%) & Peak Memory (GB) \\
\midrule

 -- & Adam & 34.01 $\pm$ 1.29  & 93.26 \\

\cmidrule(lr){1-4}
 \multirow{4}{*}{4} &
  LoRA   & 28.34 $\pm$ 2.79 & 61.70 \\
        & GaLore & 27.42 $\pm$ 1.22 & 69.00 \\
        & LDAdam & 27.93 $\pm$ 1.41 &  68.99\\
        & \cellcolor{mycolor}OASIS & \cellcolor{mycolor}29.28 $\pm$ 0.61 &  \cellcolor{mycolor}46.16\\
\cmidrule(lr){1-4}
  \multirow{4}{*}{32} &
  LoRA   & 31.84 $\pm$ 1.74 & 62.08 \\
        & GaLore & 31.24 $\pm$ 2.74 & 69.20 \\
        & LDAdam & 34.20 $\pm$ 1.43 &  69.37\\
        & \cellcolor{mycolor}OASIS & \cellcolor{mycolor}33.50 $\pm$ 2.13 & \cellcolor{mycolor}46.35 \\
\bottomrule
\end{tabular}
\caption{
Finetuning results on HumanEval for Llama-2 7B (trained on CodeAlpaca). OASIS achieves the best performance among low-rank methods at aggressive compression, and remains competitive with the strongest baseline at higher rank, while consistently using the lowest memory.
}
\label{tab:finetune_humaneval}
\end{table}

\subsection{Main Results}

\begin{table*}[htbp]
\centering
\begin{tabular}{lcc cc}
\toprule
Method 
& \multicolumn{2}{c}{Llama-130M} 
& \multicolumn{2}{c}{Llama-350M} \\
\cmidrule(lr){2-3} \cmidrule(lr){4-5}
& Val Loss & Mem 
& Val Loss & Mem \\
\midrule
Adam &  3.21 & 41.00 & 3.03 & 89.31   \\
\midrule
 GaLore & 3.55 & 40.73 & 3.39 & 88.22   \\
LDAdam & 3.23 & 40.73 & 2.86 & 88.22   \\
\cellcolor{mycolor}OASIS   & \cellcolor{mycolor}3.28 & \cellcolor{mycolor}37.10 & \cellcolor{mycolor}3.02 & \cellcolor{mycolor}79.46 \\
\bottomrule
\end{tabular}
\caption{Pretraining results on C4 across Llama model scales. OASIS achieves comparable validation loss to Adam while consistently reducing memory usage.}
\label{tab:pretraining_c4}
\end{table*}
\paragraph{Finetuning.}
Table~\ref{tab:finetune_llama_gsm8k} compares OASIS with full fine-tuning and prior low-rank methods on GSM8K. We evaluate two ranks to capture the trade-off between compression and performance: a lower rank ($r=32$) corresponding to an aggressive memory reduction regime, and a higher rank ($r=128$) chosen such that OASIS closely matches full fine-tuning performance.
For Llama-2 7B, OASIS matches the performance of full fine-tuning while reducing peak memory by nearly $2\times$. Across both ranks, it consistently outperforms prior low-rank methods. Notably, even in the low-rank regime ($r=32$), OASIS outperforms all baselines while using substantially less memory, achieving over $30\%$ lower memory usage compared to the closest performing baseline, LDAdam.
Similar trends are observed for the Llama-3.2 1B model. At a lower rank, OASIS remains competitive with prior methods while achieving the lowest memory usage, and at a higher rank, it closely approaches full fine-tuning performance. To better understand the source of memory savings, we present a breakdown of peak memory across different training components in Figure \ref{fig:memory_breakdown_baselines} (at rank $r=32$). Activation memory constitutes the largest portion of the total footprint, followed by optimizer states. LDAdam and GaLore primarily reduce optimizer memory, while LoRA reduces gradient and optimizer states' memory while leading to a larger activation footprint. In contrast, OASIS reduces memory across activations, gradients, and optimizer states, resulting in the lowest memory footprint.

Table~\ref{tab:finetune_humaneval} shows results on HumanEval for Llama-2 7B. At lower rank ($r=4$), OASIS achieves the best accuracy among all low-rank methods while using substantially less memory, highlighting its effectiveness in aggressive compression regimes.
At higher rank ($r=32$), OASIS remains competitive with the strongest baseline (LDAdam) while reducing memory usage by over $30\%$. These results further demonstrate that OASIS provides strong performance–memory trade-offs, particularly in low-rank settings.

\textbf{Pretraining.} Table~\ref{tab:pretraining_c4} shows pretraining results on the C4 dataset for Llama-130M and Llama-350M. OASIS achieves competitive validation loss compared to full training while consistently reducing memory usage across model scales.
For Llama-350M, OASIS closely matches the performance of Adam while reducing peak memory by over $10\%$. For the smaller Llama-130M model, OASIS incurs only a minor increase in validation loss while still providing meaningful memory savings.
Overall, these results demonstrate that OASIS scales effectively to pretraining settings, maintaining strong performance while reducing memory consumption.

\subsection{Ablation Study}
\begin{figure}[t]
\centering
\begin{subfigure}{0.45\linewidth}
\centering
\includegraphics[width=0.8\textwidth]{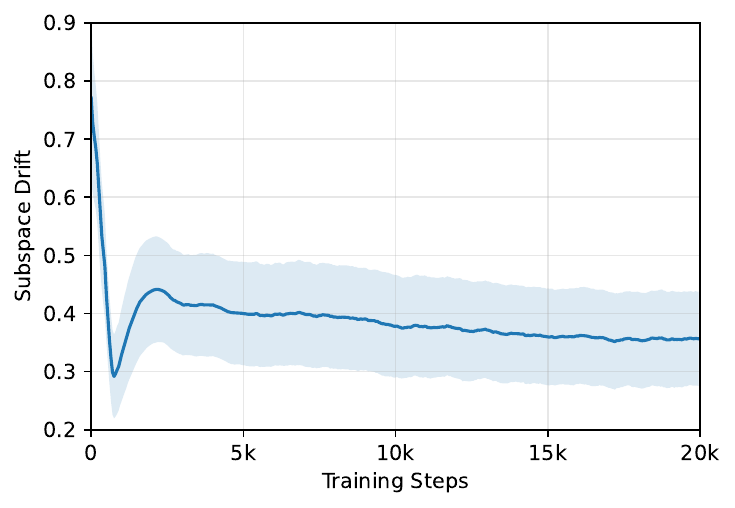}
\caption{Llama 130M}
\end{subfigure}
\hfill
\begin{subfigure}{0.45\linewidth}
\centering
\includegraphics[width=0.8\textwidth]{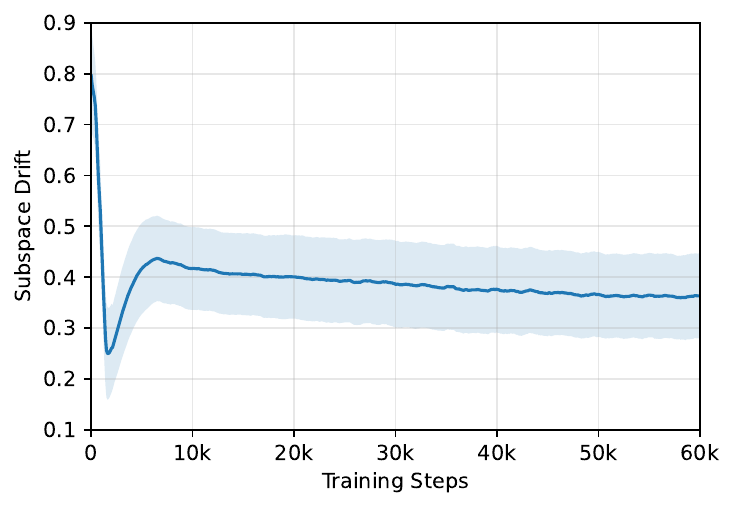}
\caption{Llama 350M}
\end{subfigure}
\caption{Subspace drift during pretraining on C4 for Llama-130M and Llama-350M. Both models exhibit a transient phase followed by stabilization at a non-zero drift, indicating that the activation subspace evolves throughout training.}
\label{fig:subtransit_pretrain}
\end{figure}

\textbf{How does the activation subspace evolve during training?}
We analyze the evolution of the activation subspace using a Frobenius-based distance between subspaces. Let $U_{t-1}, U_t \in \mathbb{R}^{d \times r}$ denote orthonormal bases of the rank-$r$ activation subspace at successive steps, and $T_t = U_t^\top U_{t-1}$ denote the subspace transition matrix. We define the subspace drift as:
\begin{equation}
\mathrm{Drift}(t) = \sqrt{1 - \frac{\|T_t \|_F^2}{r}}.
\end{equation}
This metric is bounded in $[0,1]$, where $0$ indicates identical subspaces and $1$ indicates orthogonal subspaces.
Figure~\ref{fig:intro_b} shows the subspace drift during finetuning, while Figure~\ref{fig:subtransit_pretrain} presents the corresponding behavior during pretraining on C4 with Llama-130M and LlaMA-350M. In both settings, the drift remains non-zero throughout training, indicating that the activation subspace does not converge to a fixed basis. Notably, pretraining exhibits a more pronounced transient phase: the drift decreases initially and subsequently settles at a non-zero level, indicating continued evolution of the subspace.
These observations suggest that the activation subspace is inherently non-stationary. Consequently, methods that rely on a fixed subspace estimated at initialization or infrequent updates based on batch statistics may fail to capture the evolving structure of activations, motivating the need for continuously updating the subspace.

\begin{figure}[t]
\centering
\begin{subfigure}{0.42\linewidth}
\centering
\includegraphics[width=\textwidth]{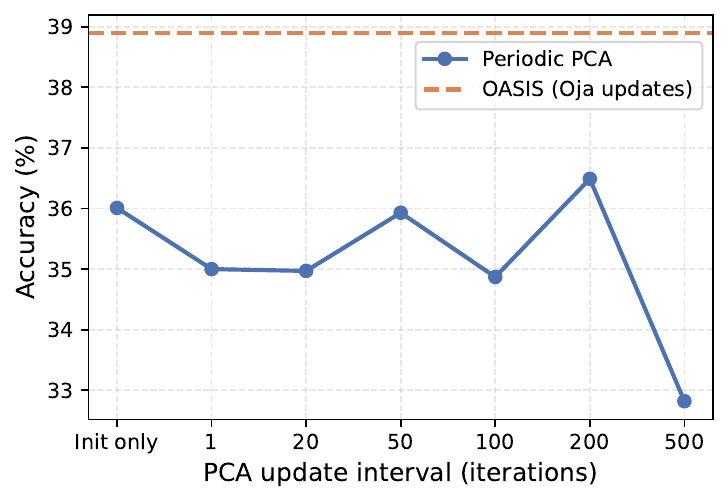}
\caption{Effect of PCA update interval}
\label{fig:pca_frequency}
\end{subfigure}
\hfill
\begin{subfigure}{0.42\linewidth}
\centering
\includegraphics[width=\textwidth]{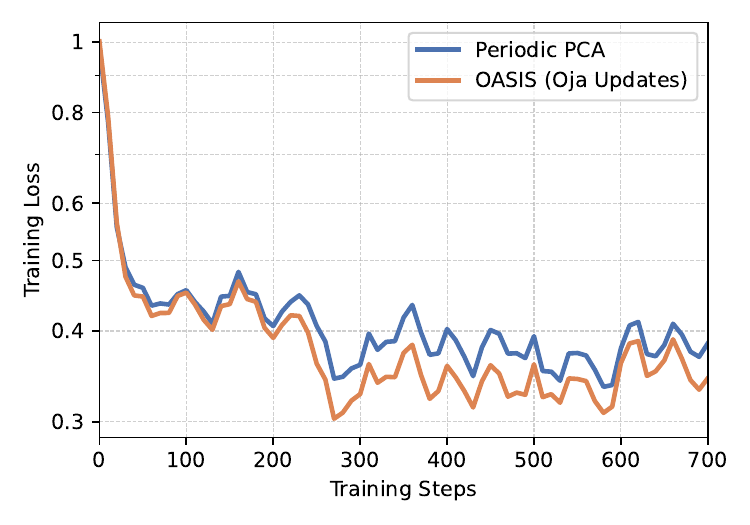}
\caption{Training loss vs iterations}
\label{fig:oja_convergence}
\end{subfigure}
\caption{Comparison of periodic PCA and OASIS during finetuning on GSM8K with Llama-2 7B. 
(a) Periodic PCA consistently underperforms OASIS across all tested update intervals.
(b) OASIS also converges faster, achieving lower training loss compared to periodic PCA.}
\label{fig:oja_ablation}
\end{figure}

\textbf{Is online subspace tracking necessary?} 
A natural question that arises is whether online subspace tracking is necessary, or if similar performance can be achieved by periodically recomputing the subspace using PCA with a carefully tuned update interval. 
To answer this, we compare OASIS with periodic PCA across a range of update intervals, where smaller intervals correspond to more frequent updates. As shown in Figure~\ref{fig:pca_frequency}, even after tuning the update interval, periodic PCA consistently underperforms OASIS.
In addition, Figure~\ref{fig:oja_convergence} shows that OASIS converges faster and achieves lower training loss compared to periodic PCA.These results highlight that relying on periodic recomputation using noisy batch statistics is fundamentally less effective than continuously tracking activation subspaces, underscoring the importance of online subspace learning for memory-efficient training.

\begin{figure}[t]
\centering
\begin{minipage}{0.44\linewidth}
    \centering
    \includegraphics[width=\linewidth]{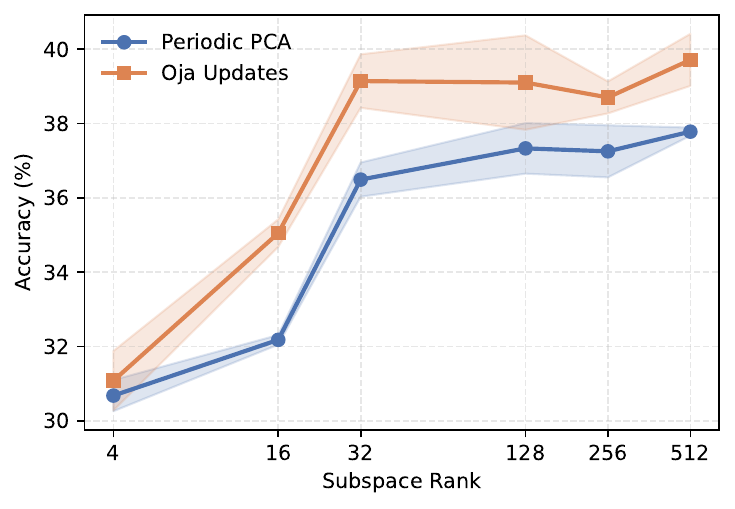}
    \caption{Effect of subspace rank on finetuning performance on GSM8K with Llama-2 7B. OASIS consistently outperforms periodic PCA across ranks.}
    \label{fig:oja_vs_pca_rank}
\end{minipage}
\hfill
\begin{minipage}{0.44\linewidth}
    \centering
    \includegraphics[width=\linewidth]{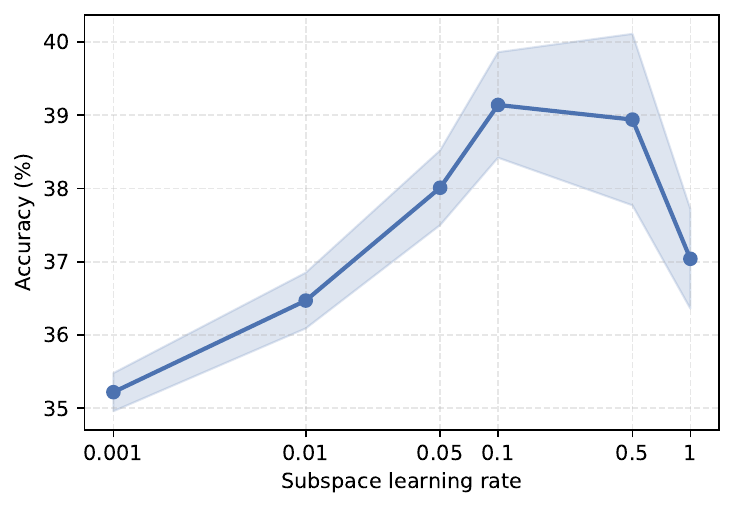}
    \caption{Effect of subspace learning rate on finetuning performance on GSM8K with Llama-2 7B. Best accuracy is achieved at 0.1 LR.}
    \label{fig:oja_lr}
\end{minipage}

\end{figure}

\textbf{How does performance vary with rank?} 
As shown in Figure~\ref{fig:oja_vs_pca_rank}, performance improves with increasing rank for both periodic PCA and OASIS, reflecting the increased expressivity of higher-dimensional subspaces. 
However, OASIS consistently outperforms periodic PCA across all ranks. Notably, OASIS achieves substantially higher accuracy even at low ranks, indicating that online subspace tracking yields more effective representations than periodic recomputation. Performance gains begin to saturate beyond moderate ranks, indicating higher-dimensional subspaces are easier to adapt to than more constrained low-rank subspaces undergoing rapid evolution.These results indicate that OASIS effectively tracks the evolving subspace and provides strong performance across a wide range of ranks.

\textbf{How does the subspace learning rate affect performance?} As shown in Figure~\ref{fig:oja_lr}, very small learning rates lead to slow adaptation of the subspace and result in lower accuracy. Increasing the learning rate improves performance, with accuracy peaking at around $0.1$. Larger learning rates degrade performance, likely due to instability in the subspace updates.
Overall, these results highlight the importance of selecting an appropriate subspace learning rate that balances stability and responsiveness in subspace adaptation. In practice, we find that a moderate value (e.g., $0.1$) consistently provides strong performance across tasks.

\section{Conclusion}
In this paper, we present OASIS, an online activation subspace learning algorithm for memory-efficient training that continuously tracks low-dimensional activation subspaces through an iterative update rule. Across both pretraining and finetuning settings, OASIS achieves strong performance while reducing memory footprint compared to prior low-rank training techniques. However, realizing these gains requires tuning the subspace learning rate (LR), which governs the dynamics of the online updates. While we observe a reasonably stable range of effective LR values, performance can degrade outside this range, making it an important hyperparameter. A promising direction for future work is to adapt this learning rate based on subspace dynamics, increasing it when the subspace is changing rapidly and decreasing it as it stabilizes.
\subsection*{Acknowledgments}
The authors would like to thank Marco Paul E. Apolinario for helpful technical discussions. This work was supported in part by, the Center for the Co-Design of Cognitive Systems (COCOSYS), a DARPA-sponsored JUMP center, the Semiconductor Research Corporation (SRC) and the National Science Foundation (NSF).
\bibliography{colm2026_conference}
\bibliographystyle{colm2026_conference}

\clearpage
\appendix
\section{Appendix}
\subsection{Extended Related Work}\label{sec:extendrelated}
\textbf{Activation Compression for Inference-Time Efficiency.}
Recent work has shown that transformer activations lie in low-dimensional subspaces and exploits this structure to improve inference efficiency in LLMs. Several approaches leverage activation subspaces for model compression and pruning, including SliceGPT \citep{slicegpt}, which reduces hidden dimensionality via subspace projection, ASVD \citep{asvd}, which performs activation-aware low-rank decomposition of model weights, and ModeGPT \citep{modegpt}, which uses activation-guided structure for pruning.
Another line of work focuses on reducing KV-cache memory by exploiting low-rank activation structure, projecting keys and values into lower-dimensional spaces during decoding \citep{eigenattn, palu, lorc, loki}. Activation-aware quantization methods further leverage this structure to compress activations with minimal accuracy loss \citep{resq}.
Collectively, these methods rely on the observation that activation covariance is approximately low-rank, enabling efficient compression, pruning, and acceleration at inference time. In contrast, we leverage activation subspace structure during training as a foundation for low-rank optimization.

\subsection{Hyperparameters}\label{apex:hyper}
\begin{table}[h]
\centering
\small
\begin{tabular}{lcc}
\toprule
\textbf{Hyperparameter} & \textbf{GSM8K (Table \ref{tab:finetune_llama_gsm8k})} & \textbf{HumanEval (Table \ref{tab:finetune_humaneval})} \\
\midrule
Batch size & 32 & 16\\
Sequence length & 512 & 1024\\
Learning rate scheduler & Cosine & Cosine\\
Learning rate (LR) & $\{2\text{e-}4, 5\text{e-}4, 9\text{e-}4, 1\text{e-}3, 1\text{e-}2\}$ & $\{2\text{e-}4, 5\text{e-}4, 9\text{e-}4, 1\text{e-}3, 1\text{e-}2\}$\\
Warmup steps & 5\% & 5\%\\
Epochs & 3 & 3\\
OASIS Subspace LR ($\gamma$) & $\{0.001, 0.01, 0.1\}$ & $\{0.001, 0.01, 0.05, 0.1\}$\\
\bottomrule
\end{tabular}
\caption{Hyperparameters for finetuning experiments on GSM8K and HumanEval.}
\label{tab:hyperparams_finetune}
\end{table}

\begin{table}[h]
\centering
\small
\begin{tabular}{lcc}
\toprule
\textbf{Hyperparameter} & \textbf{Llama-130M} & \textbf{Llama-350M} \\
\midrule
Batch size & 512 & 512 \\
Sequence length & 256 & 256 \\
Training iterations & 20k & 60k \\
Learning rate (LR) & $\{5\text{e-}4, 9\text{e-}4, 1\text{e-}3, 5\text{e-}3, 1\text{e-}2\}$ & $\{5\text{e-}4, 9\text{e-}4, 1\text{e-}3, 5\text{e-}3, 1\text{e-}2\}$\\
Learning rate scheduler & Cosine & Cosine \\
Warmup & 10\% & 10\% \\
OASIS subspace LR ($\gamma$) & $\{0.001, 0.01, 0.1\}$ & $\{0.001, 0.01, 0.1\}$ \\
\bottomrule
\end{tabular}
\caption{Hyperparameters for pretraining experiments on C4 presented in Table~\ref{tab:pretraining_c4}.}
\label{tab:hyperparams_pretrain}
\end{table}

\begin{table}[h]
\centering
\small
\begin{tabular}{lcccc}
\toprule
Model & Hidden & Intermediate & Heads & Layers\\
\midrule
Llama-130M & 768  & 2048 & 12 & 12\\
Llama-350M & 1024 & 2736 & 16 & 24\\
\bottomrule
\end{tabular}
\caption{Model configurations for pretraining experiments.}
\label{tab:llama_configs}
\end{table}

All experiments are conducted using bfloat16 (bf16) precision on NVIDIA H200 GPUs.
We summarize the hyperparameters for finetuning and pretraining experiments in Tables~\ref{tab:hyperparams_finetune} and \ref{tab:hyperparams_pretrain}, respectively. For finetuning on GSM8K and HumanEval, we use a cosine learning rate schedule with 5\% warmup and train for 3 epochs, tuning the learning rate and OASIS subspace learning rate across a fixed grid.

For pretraining on C4, we use a shared configuration across model scales, including batch size, sequence length, optimizer settings, and learning rate schedules. The primary difference lies in the number of training iterations, where LLaMA-130M and LLaMA-350M are trained for 20k and 60k steps, respectively. These iteration counts are chosen based on Chinchilla scaling laws to ensure compute-efficient training across model sizes \citep{chinchilla}. Please refer to Table~\ref{tab:llama_configs} for details on model architectures.

Hyperparameters for baseline methods are selected based on the configurations reported in their respective papers \citep{lora,galore,ldadam}.

\end{document}